\documentclass[letterpaper, 10 pt, conference]{ieeeconf}  
\IEEEoverridecommandlockouts                              
\usepackage{graphicx}
\usepackage{amsmath}
\usepackage{amssymb}
\usepackage{threeparttable}
\usepackage{booktabs}
\usepackage{color}
\usepackage{algorithm}
\usepackage{algorithmic}
\usepackage{pifont}
\usepackage{mathrsfs}
\usepackage{multirow}
\usepackage{url}
\usepackage{hyperref}
\hypersetup{hidelinks,
	colorlinks=true,
	allcolors=black,
	pdfstartview=Fit,
	breaklinks=true}

\newcommand{\cmark}{\ding{51}}
\newcommand{\xmark}{\ding{55}}

\title{\LARGE \bf
Garbage Segmentation and Attribute Analysis by Robotic Dogs
}

\author{Nuo Xu,\ Jianfeng Liao,\ Qiwei Meng\ and Wei Song\ 
\thanks{Nuo Xu, Jianfeng Liao, Qiwei Meng and Wei Song are with ZhejiangLab, Hangzhou 311100, China (e-mail:\{nuo.xu, jfliao, mengqw, weisong\}@zhejianglab.com).}%
}

\begin{document}

\maketitle
\thispagestyle{empty}
\pagestyle{empty}

\begin{abstract}
Efficient waste management and recycling heavily rely on garbage exploration and identification. In this study, we propose GSA2Seg (Garbage Segmentation and Attribute Analysis), a novel visual approach that utilizes quadruped robotic dogs as autonomous agents to address waste management and recycling challenges in diverse indoor and outdoor environments. Equipped with advanced visual perception system, including visual sensors and instance segmentators, the robotic dogs adeptly navigate their surroundings, diligently searching for common garbage items. Inspired by open-vocabulary algorithms, we introduce an innovative method for object attribute analysis. By combining garbage segmentation and attribute analysis techniques, the robotic dogs accurately determine the state of the trash, including its position and placement properties. This information enhances the robotic arm's grasping capabilities, facilitating successful garbage retrieval. Additionally, we contribute an image dataset, named GSA2D, to support evaluation. Through extensive experiments on GSA2D, this paper provides a comprehensive analysis of GSA2Seg's effectiveness. Dataset available: \href{https://www.kaggle.com/datasets/hellob/gsa2d-2024}{https://www.kaggle.com/datasets/hellob/gsa2d-2024}.
\end{abstract}

\section{Introduction}
Waste management and recycling \cite{wu2023applications} play a crucial role in maintaining a sustainable and environmentally conscious society. Effective waste management encompasses a range of activities, including waste collection, sorting, treatment, and disposal, with the ultimate goal of minimizing environmental impact and promoting resource conservation. In this context, the initial step towards these endeavors involves the exploration and identification of garbage. The utilization of robots (\textit{e.g.} quadruped robotic dogs) for waste collection presents an innovative and efficient solution \cite{fulton2019robotic}. These robotic dogs provide increased mobility and adaptability to diverse terrains, while alleviating the need for human effort. Equipped with advanced perception systems and robotic arms, they navigate both indoor and outdoor environments seamlessly. The sophisticated perception systems enable precise detection and determination of trash positions, ensuring accurate retrieval. Complemented by robotic arms, these robotic dogs effortlessly grasp and transport garbage to designated disposal bins or recycling centers, automating waste management.

In this paper, we direct our attention to the critical task of garbage segmentation \cite{sanchez2022cleansea,wang2020multi,proencca2020taco,bashkirova2022zerowaste,hong2020trashcan,koskinopoulou2021resort-it}. After the robot dog detects and navigates towards the target object, the execution of the grasping algorithm relies on the mask segmentation derived from the current observation image by mainstream instance segmentators (\textit{e.g.}, MaskRCNN \cite{he2017mask}, CascadeRCNN \cite{cai2019cascade}, MS-RCNN \cite{huang2019mask}, SOLOv2 \cite{wang2020solov2}, Mask2Former \cite{cheng2022masked}). However, this simplistic approach alone may lack the desired level of robustness. It becomes imperative to incidentally analyze the attributes of the target object during the segmentation algorithm's execution. The knowledge of object attributes, such as refined position (ground or platform), orientation (standing or lying), and deformations, significantly assists the robotic arm grasping process. These attributes aid in generating optimal grasping strategies and improving success rates. They help plan approach trajectories, adjust grip for stability, and account for object deformations. This information enhances grasp precision, reduces slippage risks, and ensures secure holds. By incorporating attribute analysis into the segmentation process, we aim to enhance the overall robustness and performance of the system.

\begin{figure}[t!]
	\centering
	\includegraphics[width=1\linewidth]{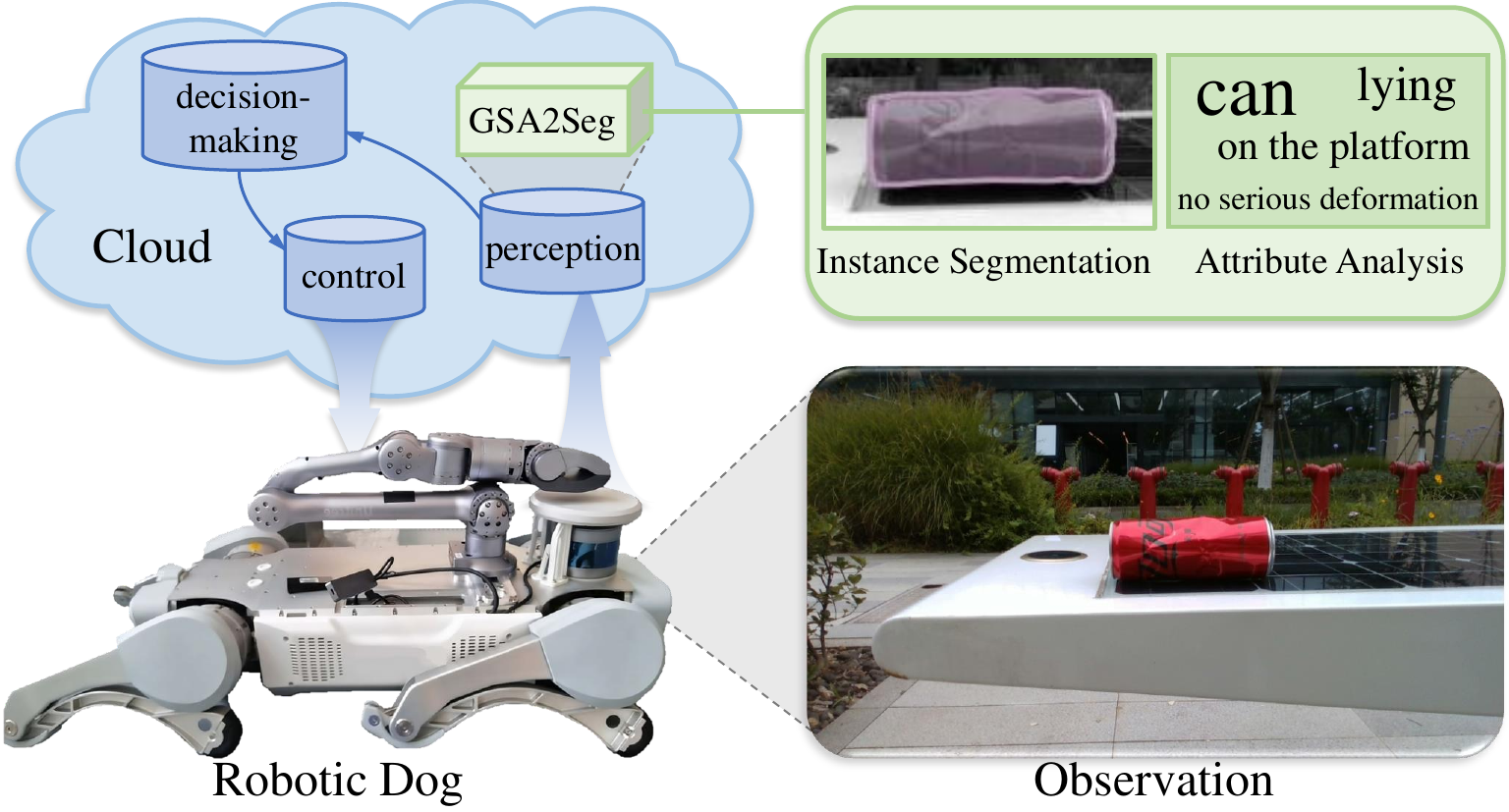}
	\caption{Garbage segmentation and attribute analysis. The robot dog utilizes visual sensors to gather environmental observation information, which is then processed by the cloud-based GSA2Seg segmenter. This cutting-edge technology enables the robot dog to perform instance segmentation and attribute analysis on garbage objects, providing accurate object masks and attributes for informed decision-making and control.}
	\label{fig:intro}
	\vspace{-0.05cm}
\end{figure}

In the context of the classic instance segmentation model, object classification lacks consideration of language-level information, treating all categories equally. While this approach may be justified in scenarios without attribute analysis, it becomes less suitable when simultaneous predictions of object categories and attributes are required. For example, a banana peel that is dropped on the ground is obviously more similar to a banana peel on the table than a can that is dropped on the ground. To overcome this limitation, we incorporate text prompts into the prediction process, enabling object categories and attributes to be treated in a similar manner as in open-vocabulary detectors. Open-vocabulary object detection \cite{radford2021learning,li2022grounded,zhou2022detecting,lin2022learning} is a fundamental task that addresses the challenge of detecting and recognizing objects from a diverse and unrestricted set of categories, without prior knowledge or constraints on object numbers or types. It poses the unique challenge of accurately localizing and classifying objects in a multi-modal way, even in the presence of novel or ambiguous instances. While these detectors are limited to predicting bounding boxes and categories, there is still valuable knowledge that can be gleaned.

Motivated by open-vocabulary detectors, we propose an innovative approach to object attribute analysis that significantly enhances the discernment of litter-related features by robotic dogs, as shown in Fig.\;\ref{fig:intro}. Our proposed algorithm, GSA2Seg, seamlessly integrates Garbage Segmentation and Attribute Analysis techniques. It enables robot dogs to precisely determine the segmentation mask and state of garbage, including its precise location and placement attributes. On the other hand, during our extensive research, we discovered a lack of attribute annotations in almost all publicly available garbage segmentation datasets. In response, we contribute GSA2D, an image dataset that supports the evaluation and benchmarking of our proposed method. Through exhaustive exploration of GSA2Seg's performance on GSA2D, combined with a comprehensive analysis of its effectiveness, we demonstrate the remarkable capabilities and potential applications of our proposed approach. The key contributions of this paper can be summarized as follows:

\vspace{-0.03cm}
\begin{itemize}
	\item Motivated by open-vocabulary algorithms, we propose GSA2Seg for garbage segmentation and attribute analysis, which enables precise determination of both object locations and attributes.
	\item To support the evaluation and benchmarking of our proposed approach, we contribute GSA2D. This dataset contains a diverse collection of images annotated with segmentation masks and attribute labels.
	\item Through extensive experimentation and analysis, we present comprehensive results and evaluation metrics that showcase the effectiveness of GSA2Seg.
\end{itemize}

\section{Methodology}
\subsection{Overview}
\label{sec:overview}

As depicted in Fig.\;\ref{fig:pipeline}, the robot dog perceives visual signals using the depth camera mounted on its head, transmitting this information to the cloud-based GSA2Seg framework. GSA2Seg comprises an encoder and a decoder. The encoder processes the visual signal and the language prompt independently and fuses them using a bi-directional attention mechanism before sending them to the decoder. Within the decoder, randomly initialized object queries are combined with visual features, leveraging attention mechanisms to predict masks and bounding boxes accurately. Furthermore, these features are attentively combined with language features for comprehensive attribute analysis, enhancing the robot dog's understanding of the objects in its environment. The integration of these components enables the robot dog to efficiently perform instance segmentation and attribute analysis for effective decision-making and control.

\subsection{Visual Perception System}
\label{sec:garbage}
The robot dog's visual perception system comprises three key components: visual sensors, instance segmenter, and attribute analyzer. Vision sensors are strategically positioned on the robot dog to capture and process visual signals from the environment. The instance segmenter and attribute analyzer, which share the same backbone architecture, form the GSA2Seg model deployed on the cloud. This integrated approach enables the robot dog to effectively explore, segment, and analyze the attributes of detected objects, facilitating precise and informed decision-making in waste management.

\noindent \textbf{Visual Sensors.}\;
The robot dog's head is equipped with two advanced depth cameras from Intel: the RealSense 455, tilted up 15°, and the RealSense 435i, tilted down 45°. The D455 is utilized for exploring visual information beyond 3m, while the D435i focuses on observing close-ups within 3m. These cameras are positioned at approximately 0.5m above the ground. Both cameras, part of the RealSense product line for robotics, offer precise RGB and depth sensing at a resolution of 1280x720, making them highly versatile for various applications. Notably, in this work, we exclusively utilize RGB images and do not incorporate depth images.

\begin{figure*}[t!]
	\centering
	\includegraphics[width=1\linewidth]{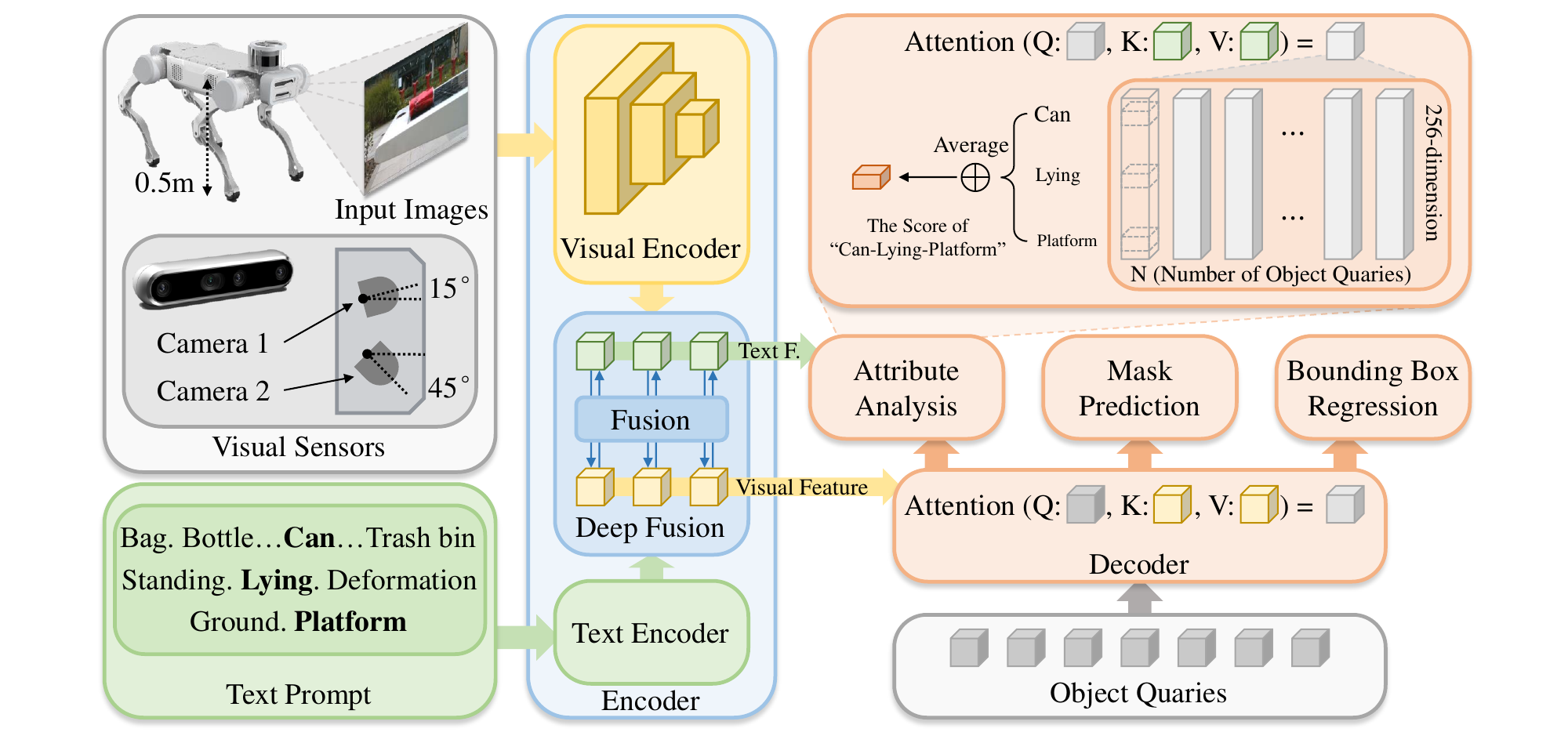}
	\caption{Pipeline of our GSA2Seg method. The robot dog utilizes its depth camera on the head to perceive visual signals, which are then transmitted to the cloud-based GSA2Seg system. GSA2Seg comprises an encoder and a decoder. The encoder processes the visual signals and language prompts separately and fuses them using a bi-directional attention mechanism before sending them to the decoder. Within the decoder, randomly initialized object queries are combined with visual features through attention mechanisms and used for predicting masks and bounding boxes. Additionally, these merged features are attentively combined with language features to facilitate attribute analysis.}
	\label{fig:pipeline}
	\vspace{-0.0cm}
\end{figure*}

\begin{table*}[th]
	\centering
	\small
	\caption{Garbage segmentation datasets comparison.}
	\vspace{-0.1cm}
	\label{tab:dataset}
	\setlength{\tabcolsep}{3.35mm}{
		\begin{tabular}{l|ccccccc}
			\toprule
			Dataset (Seg.) & No. categories & No. images & Resolution & Sensors & No. annotations & Attribute & Comment  \\
			\midrule
			\midrule
			CleanSea \cite{sanchez2022cleansea} & 19 & 1.2k & 640$\times$480 & RGB & 2.0k & \xmark & underwater  \\
			MJU-Waste \cite{wang2020multi} & 1 & 2.5k & 640$\times$480 & RGB-D & 2.5k & \xmark & indoor \\
			TACO \cite{proencca2020taco} & 28 & 1.5k & 3223$\times$2825 & RGB & 4.8k & \xmark & outdoor \\
			\midrule
			\midrule
			\textit{Ours} GSA2D & 10 & 3.1k & 1024$\times$720 & RGB-D & 8.6k & \cmark & in/outdoor \\
			\bottomrule
	\end{tabular}}
	\vspace{-0.27cm}
\end{table*}

\noindent \textbf{Garbage Segmentor.}\;
Upon receiving the visual perception signal from the robot dog, GSA2Seg, deployed in the cloud, performs instance segmentation on the current image. GSA2Seg consists of two main components: the encoder and the decoder. The encoder follows the approach presented in \cite{li2022grounded} and incorporates a two-way attention mechanism, allowing interactive fusion of visual information and language prompts. For the decoder, we made enhancements solely to the classification header, based on the model proposed in \cite{yan2023universal}, while retaining the original architecture design for the remaining components. Since the technology used in the garbage segmentation module is well-established, we opted for a classic model, enabling us to concentrate on the subsequent attribute analysis tasks. This strategic choice allows us to focus our efforts on advancing attribute analysis, building upon the established foundation of the instance segmentation module.

\noindent \textbf{Attribute Analyzer.}\;
A key improvement of GSA2Seg lies in its ability to predict various attributes of the target object. In a standard decoder \cite{cheng2022masked}, Object Queries are typically used to query specific regions in the image by attending to visual features. This allows the generation of masks, bounding box predictions, and object classification. However, in this form of object classification, language-level information is not taken into account, treating each category equally. While this approach may be reasonable without considering attribute analysis, it becomes less appropriate when simultaneous prediction of object categories and attributes is required. To address this limitation, we introduce language prompts and utilize language features to perform additional attention in conjunction with fusion features, enabling object attribute analysis. As depicted in Fig.\;\ref{fig:pipeline}, the Object Query feature, which undergoes two rounds of fusion with visual and language features, is utilized for object attribute analysis. Each dimension of the final fusion feature represents a specific category or attribute, and the combination of these features allows us to calculate the average value, yielding the prediction score for the corresponding attribute combination. During training, the scores obtained through this combination replace the original category scores for loss calculation, ensuring the incorporation of attribute analysis into the learning process. This novel approach enhances the accuracy and robustness of the object attribute prediction, capturing subtle differences between objects of the same category based on their attributes. By incorporating language-level information and performing attentive attribute analysis, GSA2Seg achieves meaningful object attribute predictions. 

\subsection{Dataset Construction}
\label{sec:dataset}
This chapter provides an overview of the current landscape of garbage instance segmentation datasets and outlines the creation process of our novel dataset, Garbage Segmentation and Attribute Analysis Dataset (GSA2D). In contrast to existing datasets, our GSA2D goes beyond instance segmentation mask annotations by also providing detailed attribute annotations for each piece of garbage.

\noindent \textbf{Existing Datasets.}\; For tasks like robotic grasping, relying solely on bounding boxes for object recognition may not be sufficient in capturing intricate object shapes and edges. This highlights the increasing importance of instance segmentation. However, the availability of discarded image datasets specifically designed for segmentation remains limited compared to other types of datasets. For instance, Tab.\;\ref{tab:dataset} showcases datasets like TACO \cite{proencca2020taco}, an open-access crowdsourced dataset with high-resolution images of littering captured by mobile phones in various real-world environments. Another dataset, MJU-Waste \cite{wang2020multi}, pioneers garbage segmentation with RGB-D images captured by researchers holding garbage in front of a depth camera. Additionally, CleanSea \cite{sanchez2022cleansea}, assembled from the JAMSTEC electronic library of deep-sea images, serves as a benchmark corpus of 1.2k images of underwater debris. This dataset includes bounding boxes and pixel masks, encompassing samples of animals, plants, and other litter categories collected from the seas surrounding Japan. In the domain of intelligent garbage recognition, segmentation primarily refers to instance segmentation, where the objective is to predict the class label for each pixel in an image, treating multiple objects of the same class as distinct individual instances. Consequently, labeling segmented data becomes a time-consuming task, potentially contributing to the scarcity of corresponding datasets. To address these limitations and cater to the specific demands of object recognition tasks, we present a new and comprehensive garbage segmentation and attribute analysis dataset named GSA2D.

\noindent \textbf{Data Collection.}\; The robot dog adeptly traverses diverse indoor and outdoor environments, capturing images featuring garbage within an approximate 8-meter range using its two head-mounted depth cameras. These images are categorized based on their proximity to the garbage, creating two groups: distant views for the robot dog's visual perception training during the exploration stage, and close shots for the visual perception training during the grasping stage. To strike a balance between data distribution and object scale fitting, the number of distant views is approximately twice that of the close shots. This decision is influenced by the smaller object scale in the exploration phase, necessitating a greater amount of data for accurate fitting. The dataset comprises a total of six distinct scenes, with five indoor scenes (including two tea rooms, two meeting rooms, and one laboratory) and one outdoor scene. The diversity of scenes offers a comprehensive representation of real-world environments, enriching the dataset with varied lighting conditions, backgrounds, and potential occlusions. The inclusion of both indoor and outdoor scenes ensures the robot dog's adaptability and robustness in handling different settings, contributing to the dataset's relevance and applicability to real-world scenarios.

\noindent \textbf{Annotation Form.}\; The GSA2D dataset comprises a total of 3119 images, featuring ten common types of garbage targets, including bottle, cup, box, can, paper ball, bag, peel, toy, cigarette, and trash bin. Each object is meticulously annotated at the instance segmentation level, providing both rectangular bounding boxes and precise polygon masks for accurate localization. Notably, four specific object types (bottle, cup, box, can) are further classified based on their state, specifically standing, lying, or deformation, represented by labels 0, 1, and 2, respectively. Deformation is treated as a separate category due to its distinct characteristics, making it challenging to determine whether the object should be considered as standing or lying down. Furthermore, the dataset includes information about the position of each target, distinguishing between objects placed on the ground (denoted by 0) and those situated on a platform (denoted by 1). This attribute annotation provides valuable insights into the objects' spatial arrangement, enhancing the dataset's utility for various tasks, such as robotic grasping and waste management. Overall, the comprehensive annotations in GSA2D enable robust analysis and accurate attribute-based object recognition, fostering advancements in intelligent robotic waste management systems.

\section{Experiments}

\subsection{Experimental Setup}
\label{sec:setup}

\noindent \textbf{Baselines.}\; We compare the proposed method GSA2Seg with five classic instance segmentation techniques as baselines (MaskRCNN \cite{he2017mask}, CascadeRCNN \cite{cai2019cascade}, MS-RCNN \cite{huang2019mask}, SOLOv2 \cite{wang2020solov2}, and Mask2Former \cite{cheng2022masked}). Notably, these traditional methods lack the ability to learn category labels as text and can only map features to distinct neurons for classification. To ensure a fair comparison, the attributes in the GSA2D dataset are combined with each of the ten categories, resulting in a total of 36 subdivision categories. Specifically, the bottle, cup, box, and can categories have three state attributes and two location attributes simultaneously, while the remaining six types are distinguished solely based on whether they are positioned on the ground or on a platform. This meticulous categorization in the dataset facilitates a comprehensive evaluation of GSA2Seg's performance and its ability to handle a wide range of attributes, making it an essential benchmark for robust instance segmentation and attribute analysis in the context of garbage recognition.

\noindent \textbf{Settings.}\; To ensure a rigorous and unbiased evaluation of various methods, we conducted experiments using the widely recognized MMDetection \cite{mmdetection} platform. MMDetection is a comprehensive repository of object detection and instance segmentation algorithms, encompassing a wide array of well-known approaches. All models are trained under a uniform framework, with an initial learning rate of 0.02 and an SGD optimizer. The maximum epoch is set to 48, while all other settings adhere to the default configurations in MMDetection. For our experiments, we randomly divide the GSA2D dataset into a training set comprising 2495 images and a validation set comprising 624 images. The models are trained on the training set and subsequently evaluate and report on the validation set. This careful setup ensures reliable comparisons and meaningful insights into the performance of GSA2Seg in relation to other instance segmentation methods.

\noindent \textbf{Evaluations.}\; GSA2D serves as a comprehensive benchmark for instance segmentation and attribute analysis tasks, featuring three essential evaluation metrics: $AP$ (Average Precision), $AP_{50}$ ($AP$ at IoU 0.5), and $FPS$ (Frames Per Second). $AP$ calculates the average precision across various IoU thresholds, ranging from 0.5 to 0.95 with a step size of 0.05, offering a holistic assessment of instance segmentation performance. $AP_{50}$ specifically focuses on precision when there is at least a 50\% overlap between predicted and ground truth masks. Furthermore, $FPS$ is employed to gauge the model's inference speed. These metrics provide crucial insights into the accuracy and efficiency of the models. Higher scores in these metrics signify superior performance in the instance segmentation and attribute analysis tasks.

\subsection{Performance Analysis}
\label{sec:performance}
\begin{figure*}[t!]
	\centering
	\includegraphics[width=1\linewidth]{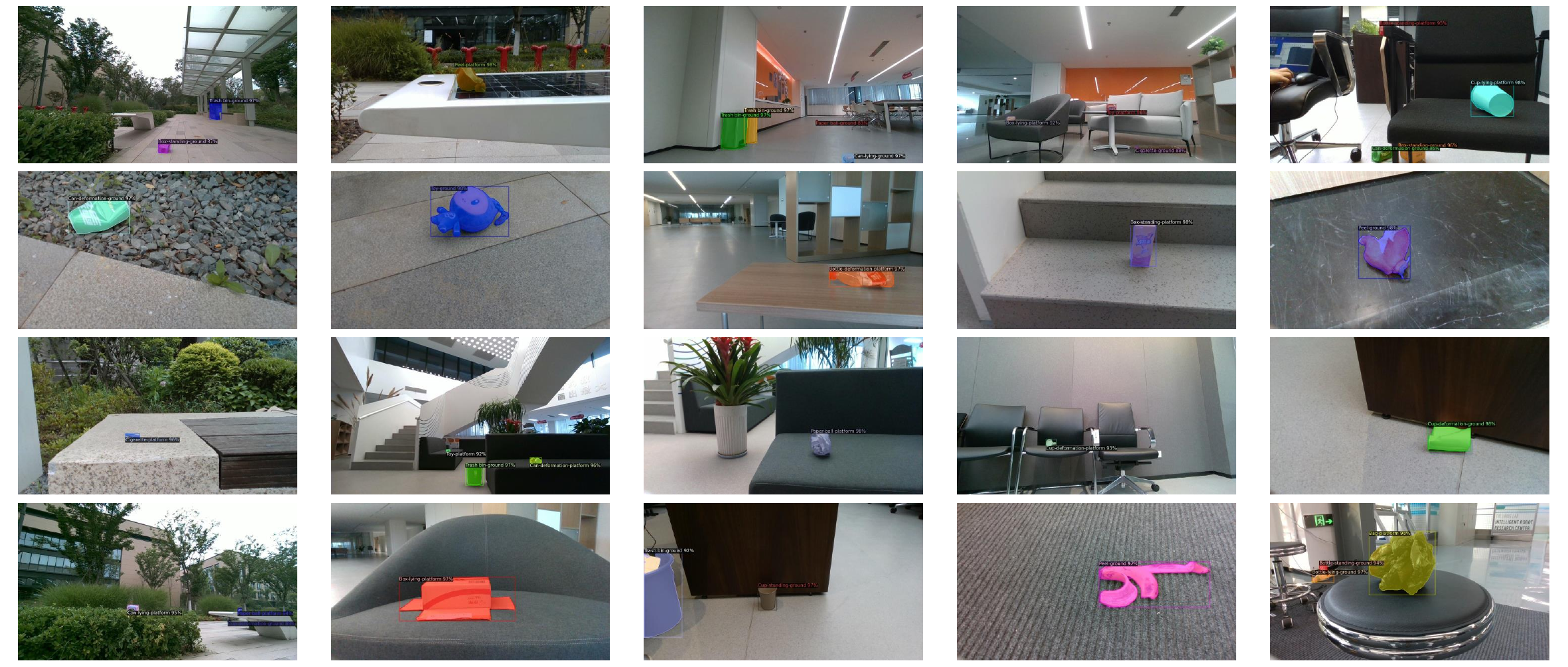}
	\caption{Qualitative Results (zoom in for detailed viewing). The figure presents a subset of our visualization results, showcasing images captured from diverse perspectives in both indoor and outdoor environments using two cameras. Among them, the predicted attributes and categories of detected garbage are marked as labels above the masks and the bounding boxes.}
	\label{fig:qualitative}
	\vspace{-0.1cm}
\end{figure*}

In this section, we will conduct both quantitative and qualitative analyses of Tab.\;\ref{tab:performance} and Fig.\;\ref{fig:qualitative} to thoroughly assess the effectiveness of GSA2Seg. Through these analyses, we aim to gain a comprehensive understanding of the performance and capabilities of our proposed approach.

\noindent \textbf{Quantitative Results.}\; Tab.\;\ref{tab:performance} presents a comprehensive comparison of performance among several classic instance segmentation models and GSA2Seg. To ensure fairness, all models employ ResNet-50 as the backbone architecture. Remarkably, GSA2Seg exhibits notable advantages in performance metrics, achieving the highest $AP$ and $AP50$ at 54.2\% and 76.3\%, respectively. While SOLOv2 boasts a fast inference speed of 31.3 $FPS$, its accuracy rate is comparatively low. On the other hand, Mask2Former ranks second in performance, but suffers from a slower inference speed. In contrast, GSA2Seg strikes a balance between inference speed and detection accuracy, making it a compelling choice for practical applications. The remarkable performance of GSA2Seg can be attributed to its utilization of multi-modal features for attribute analysis. Unlike classic segmenters that rely on selecting neurons to distinguish subdivision labels of the fusion category and attribute information, GSA2Seg employs natural language to achieve cross-modal matching in the multimodal feature space. This innovative approach enhances the model's ability to understand and interpret complex visual scenes, resulting in superior performance in attribute analysis and instance segmentation tasks.

\noindent \textbf{Qualitative Results.}\; Fig.\;\ref{fig:qualitative} presents a visual representation of the instance segmentation and attribute analysis results obtained through GSA2Seg. The images depict diverse scenes, ranging from outdoor to indoor environments, and encompassing various perspectives, from distant to close-up views. The results showcase ten different types of litter objects placed on either the ground or platforms at varying heights. Moreover, the visualization highlights different placement states of instances within categories like bottles, boxes, cans, and cups. This compelling visualization serves to demonstrate the remarkable effectiveness of GSA2Seg in precisely identifying and analyzing waste objects across a wide array of scenarios and attributes.

\begin{table}[t]
	\centering
	\small
	\caption{Garbage segmentation performance comparison.}
	\vspace{-0.1cm}
	\label{tab:performance}
	\setlength{\tabcolsep}{5.0mm}{
		\begin{tabular}{l|ccc}
			\toprule
			Model (ResNet-50)  & $AP$ & $AP_{50}$ & $FPS$ \\
			\midrule
			\midrule
			SOLOv2 \cite{wang2020solov2}  & 29.9 & 45.0 & \textbf{31.3} \\
			MaskRCNN \cite{he2017mask} & 51.0 & 70.7 & 16.1 \\
			CascadeRCNN \cite{cai2019cascade} & 51.4 & 70.9 & 11.2 \\
			MS-RCNN \cite{huang2019mask}  & 52.0 & 69.0 & 11.8 \\
			Mask2Former \cite{cheng2022masked}  & 53.2 & 75.6 & 8.6 \\
			\midrule
			\midrule
			\textit{Ours} GSA2Seg  & \textbf{54.2} & \textbf{76.3} & 10.7 \\
			\bottomrule
	\end{tabular}}
	\vspace{-0.0cm}
\end{table}

\begin{table}[t]
	\centering
	\small
	\caption{Ablation study on model component decomposition.}
	\vspace{-0.1cm}
	\label{tab:components}
	\setlength{\tabcolsep}{3.4mm}{
		\begin{tabular}{cccc|cc}
			\toprule
			prompt  & bbox & mask & attribute & $AP$ & $AP_{50}$ \\
			\midrule
			\xmark & \cmark & \cmark & \cmark & 51.9 & 71.4 \\
			\cmark & \cmark & \xmark & \xmark & 49.2 & \textbf{79.4} \\
			\cmark & \cmark & \cmark & \xmark & 53.5 & 79.0 \\
			\cmark & \cmark & \cmark & \cmark & \textbf{54.2} & 76.3 \\
			\bottomrule
	\end{tabular}}
	\vspace{-0.37cm}
\end{table}

\subsection{Ablation Study}
\label{sec:ablation}
In this section, we present the results of three ablation studies to analyze the contributions of different components in the model GSA2Seg and different categories and attributes in the dataset GSA2D to the overall performance, by removing or modifying specific modules.

\noindent \textbf{Model Components.}\; As depicted in Tab.\;\ref{tab:components}, our ablation experiments primarily focus on the head of GSA2Seg. When considering only visual information in the attribute analysis header without fusing the text prompt, we observe a noticeable drop of 2.3\% in $AP$, indicating that the addition of language information significantly enhances performance. Furthermore, if the model solely performs simple 10-classification without involving complex attribute analysis, the $AP_{50}$ index, describing loose segmentation performance, increases by 2.7\%, while the stricter comprehensive $AP$ index decreases by 0.7\%. Notably, if we omit instance segmentation and solely perform bounding box detection, this trend becomes more pronounced. This phenomenon may be attributed to the model's ease of learning with simpler annotation information but its inability to achieve more sophisticated and accurate results. These ablation results provide valuable insights into the crucial components that contribute to GSA2Seg's robust performance in garbage segmentation and attribute analysis.

\noindent \textbf{Each Category.}\; As depicted in Tab.\;\ref{tab:category}, our primary focus in the ablation experiments lies on the all 10 categories of GSA2Seg in GSA2D. Among these categories, the trash can demonstrates the highest segmentation performance, achieving an impressive $AP$ value of 71.6\%. This is likely due to the relatively regular appearance and distinct characteristics of trash cans in the collected images. On the other hand, cigarettes pose the most challenging garbage type to identify, as their size is often small and they occupy only a limited number of pixels in the images. Consequently, cigarettes achieve a lower $AP$ value of 32.9\%. These findings shed light on the varying levels of difficulty associated with different garbage categories in the GSA2D dataset.

\noindent \textbf{Each Attribute.}\; Tab.\;\ref{tab:category} further provides insights into the performance under different attributes in GSA2D. Overall, the recognition performance across the five attributes tends to be relatively consistent. For instance, the $AP$ values for "on the ground" and "on the platform" are 52.2\% and 56.2\%, respectively, showing little variation. However, it is notable that identifying object deformations proves to be more challenging than determining whether the object is in a standing or lying position. This could be attributed to the model's limitations in understanding three-dimensional information, which presents an area for potential improvement in future iterations. Addressing this aspect could enhance the overall robustness and accuracy of GSA2Seg in analyzing attributes.

\section{Conclusions}
In conclusion, our study presents GSA2Seg, an innovative visual approach for efficient garbage segmentation and attribute analysis. Equipped with advanced perception systems, the robotic dogs adeptly navigate diverse environments, successfully identifying common garbage items and analyze their attributes for improved robotic arm grasping. Additionally, we contribute the GSA2D dataset, facilitating evaluation and benchmarking. Our experiments demonstrate the effectiveness of GSA2Seg in waste management applications, paving the way for automated garbage identification. The seamless integration of visual perception, attribute analysis, and robotic control holds promise for driving sustainable practices in waste management and recycling.

\begin{table*}[th]
	\centering
	\small
	\caption{Ablation studies on categories and attributes.}
	\vspace{-0.1cm}
	\label{tab:category}
	\setlength{\tabcolsep}{8.4mm}{
		\begin{tabular}{l|r||l|r||l|r}
			\toprule
			Each category & $AP$ & Each category & $AP$ & Each attribute & $AP$ \\
			\midrule
			C01-Bag & 58.8 & C06-Cup & 62.2 & On the ground & 52.2 \\
			C02-Bottle & 50.9 & C07-Paper ball & 48.8 & On the platform & 56.2 \\
			C03-Box & 51.5 & C08-Peel & 51.9 & Standing & 59.2 \\
			C04-Can & 54.3 & C09-Toy & 54.7 & Lying & 57.6 \\
			C05-Cigarette & 32.9 & C10-Trash bin & 71.6 & Deformation & 47.4 \\
			\bottomrule
	\end{tabular}}
	\vspace{-0.2cm}
\end{table*}

\bibliographystyle{IEEEtran}
\bibliography{paper}

\end{document}